# Language-based Examples in the Statistics Classroom

Roger Bilisoly[1]

[1]Department of Mathematical Sciences, Central Connecticut State University,
1615 Stanley St, New Britain, CT 06050-4010

**Abstract**
Statistics pedagogy values using a variety of examples. Thanks to text resources on the Web, and since statistical packages have the ability to analyze string data, it is now easy to use language-based examples in a statistics class. Three such examples are discussed here. First, many types of wordplay (e.g., crosswords and hangman) involve finding words with letters that satisfy a certain pattern. Second, linguistics has shown that idiomatic pairs of words often appear together more frequently than chance. For example, in the Brown Corpus, this is true of the phrasal verb *to throw up* (p-value=7.92E-10.) Third, a pangram contains all the letters of the alphabet at least once. These are searched for in Charles Dickens' *A Christmas Carol*, and their lengths are compared to the expected value given by the unequal probability coupon collector's problem as well as simulations.

**Key Words:** Linguistics, Wordplay, Fisher's exact test, coupon collector's problem

## 1. Analyzing Language

Although analyzing language data is uncommon among statisticians, there are many linguists that apply statistical techniques. For example, corpus linguists collect language samples that are representative of a particular aspect of a language, which are then analyzed using tools from both statistics and information theory. The Brown Corpus, for instance, was created to be representative of American English in 1961 and consists of 500 samples each containing about 2000 words. Computational linguists are also sophisticated users of statistics. In fact, there are statistically orientated books written by linguists such as Manning and Schütze (1999) and Oakes (1998). Hence, if a statistics teacher wishes to use text data in class, many examples have already been worked out.

This paper gives three examples of applying statistics to language. The first applies a string pattern matching methodology called regular expressions to wordplay. The second discusses *collocations*, which is a concept from linguistics. Finally, the third applies more sophisticated mathematical techniques to pangrams, which is a type of wordplay. These examples should give the reader a taste of what can be done with language. But beware, once one starts it can be hard to stop.

## 2. Wordplay and Regular Expressions

Several types of wordplay and word games require finding a word with letters satisfying a pre-specified string pattern. These can be formed by using a programming methodology called *regular expressions*, or *regexes* for short. It is useful because it is implemented in a variety of packages. In this paper we will use SAS's implementation of Perl regular expressions, which first appeared in version 9. Only some simpler patterns are shown here: for a more thorough introduction to regexes and their use in text mining, see chapter 2 of Bilisoly (2008).

## 2.1 Crossword Puzzle Example

We first consider crossword puzzles. Here the length of an answer is known, and if the puzzle is partially finished, then some of the letters in the answer may be known. For example, consider a seven letter word with a *b* in the fourth position and a *u* in the last position. How informative is this, which is to say, how many words satisfy these constraints? This can be easily done in two steps. First, read a wordlist into SAS as shown in Figure 1. The file `crosswd.txt` is from Ward (2002), which is freely available from Project Gutenberg. This wordlist is used for all the code samples in this section.

```
data wordlist;
length word $30.;
infile "c:\crosswd.txt";
input word $; run;
```

**Figure 1:** SAS code to read in one of Grady Ward's Moby Word Lists.

Using the dataset `wordlist` created by the SAS code in Figure 1, Figure 2 uses a regular expression called `regex` to select the seven letter words that satisfy the crossword constraints given above. The forward slashes are delimiters of the regex, and the period stands for any one character. The symbol ^ stands for the beginning of the line, which is the beginning of the word since `crosswd.txt` contains exactly one word per line. Note that SAS likes fixed length string variables, which causes blanks to be appended to the words, hence the need for a blank after the letter `u`. The + stands for one or more of the character preceding it (in this case one or more blanks), and the $ stands for the end of the line.

```
data crossword; set wordlist;
keep word;
if _n_ = 1 then regex = prxparse("/^...b..u +$/");
retain regex;
start_match = prxmatch(regex, word);
if start_match > 0 then output; run;

proc print data=crossword; run;
```

**Figure 2:** SAS code to find all seven letter words such that the fourth letter is *b* and the last letter is *u*.

Figure 3 gives the results of running the SAS code in Figures 1 and 2. It turns out that exactly one word satisfies the conditions imposed. Hence in this case, the letter pattern is quite informative.

```
Obs     word

 1     jambeau
```

**Figure 3:** SAS output after running the code given in Figures 1 and 2. It turns out that there is only one seven letter word such that the fourth letter is *b* and the last letter is *u*.

Of course, this is not the only way to solve this problem. The same result can be obtained by using the string functions `substr()` and `length()`. However, using regular expressions has the advantage of portability to many other software packages and programming languages. See Friedl (2006) for more on comparing implementations of regexes.

### 2.2 Hangman Example

Hangman is somewhat like a single word in a crossword puzzle except that there can be letters known not to be in the word. Also, if a guessed letter appears in a word, all instances of it are revealed. These new constraints are easily implemented using regular expressions. For example, suppose a seven letter word has *e* for the second letter and ends in *s* (so none of the unknown letters are either *e* or *s*), plus the letters *t*, *a*, *o*, *i*, *n*, have been guessed but do not appear in this word. Figure 4 gives SAS code that can find all such words.

```
data hangman; set wordlist;
keep word;
if _n_ = 1 then regex =
prxparse("/^[^etaoins]e[^etaoins]{4}s +$/");
retain regex;
start_match = prxmatch(regex, word);
if start_match > 0 then output; run;

proc print data=hangman(obs=30); run;
```

**Figure 4:** SAS code to find all seven letter words such that the second letter is *e* and the last letter is *s*, plus the letters *t*, *a*, *o*, *i*, *n*, do not appear and none of the unknown letters are either *e* or *s*.

Note that Figure 4 is almost exactly the same as Figure 2: the only differences are the regular expression used. Note that square brackets starting with a ^ means not to match any of the bracketed letters. Hence the ^ has a different meaning inside square brackets than outside them. Finally, the {4} means four characters satisfying the letter restriction immediately preceding it. Running the SAS code in Figures 1 and 4 produces Figure 5, which shows twelve possible words.

```
Obs     word

 1     bedbugs
 2     bedrugs
 3     bedumbs
 4     begulfs
 5     ferrums
 6     peplums
 7     rebuffs
 8     redbuds
 9     redbugs
10     regulus
11     vellums
12     zephyrs
```

**Figure 5:** SAS output after running the code given in Figures 1 and 4. It turns out that there are twelve seven letter words such that the second letter is *e* and the last letter is *s*, plus the letters *t*, *a*, *o*, *i*, *n*, do not appear and *e* or *s* do not appear twice.

Many other word games or types of wordplay involve finding words that satisfy some pattern. Once one learns regular expressions, however, many of these are easy to find by just placing the appropriate regex into the Figure 2 code. For example, what is the longest word in English? Answer: the word *smiles* because it is a mile between the first and last letter. Are there longer words that contain the substring *mile*? Yes, there are, and it is left as a SAS programming exercise to find all of these.

### 3. Word Collocations

The preceding section provides two examples that involve wordplay, and these interest students who enjoy such recreational activities. However, word analyses can also serve a more serious purpose in linguistics. The example of word collocations is given in this section, and it is both important to corpus linguists and illustrates categorical data analysis.

**3.1 Examples of Word Collocations**
In linguistics, word collocations are two (or more) words that appear as a unit, which has a meaning that is not obvious from the meanings of the constituent words. For example, "white house" could mean just an arbitrary house that is white, but it appears frequently in print because that is where the president of the United States lives. This latter definition, however, cannot be deduced by knowing the definitions of the words "white" and "house." Hence, "white house" is an example of a word collocation.

One way to find word collocations is to compare the frequency of the constituent words to the frequency of the words together. Positive correlation suggests a collocation, while independence does not. Looking at English texts, one can also find negatively correlated words, e.g., "ultraviolet house," which only appears 110 times when searched for on Google (on 9/16/2009). Since no house is ultraviolet in color, this is not surprising.

Word collocations have been traditionally found by concordancing, which is a sorted list of word matches. Not surprisingly, regular expressions are useful here. Figure 6 shows partial results of searching the Brown Corpus for the word *up*. Here the words to the left of *up* have been sorted, which is useful for finding collocations. This output was produced by Program 6.1 from Bilisoly (2008).

```
 1 bringing the level of acquaintance up to adequacy for future cooperative
 2 day. Hubie's restaurant activities up in Lorain, Ohio, may preclude his
 3  December 31? What does it all add up to? Indications are that Khrushche
 4 ing ( most of it inefficient ) add up to one-fourth of the total constru
 5  2,500 such projects, and they add up to a lot more than just roads and
 6 ne chuckled. How often do they add up to headlines? You should complain.
 7 and each reading a score by adding up these weights. Specific dates woul
 8 totted up and tabulated, by adding up the Hits and Significants, with th
 9 an condition  --  the whole adding up to nothing more than a glimpse int
10 ale is very high. Even so, it adds up to impossible odds, except that th
11 Kirov for the time being. It's all up in the air again. So the Kirov wil
12 , as I write, being fought out all up and down those streets. Northerner
```

**Figure 6:** A few instances of the word *up* in the Brown Corpus, where the lines are sorted by the word to the left of *up*. Program 6.1 of Bilisoly (2008) produced this output.

Before moving to the next section, note that concordancing has many uses in linguistics, two of which are explained here. First, word collocations are useful in finding idioms, and these are essential for language teachers because, by definition, students who know the individual words may not know what the collocation means. For example, in Figure 6, one sees that the preposition *up* of the phrasal verb *to add up* has no relation to the common meaning of *up* (referring to a higher position). In fact, saying "add together this list of numbers" makes more sense, but that is not how it is said in English. However, German, in fact, does say it that way with the separable verb *zusammenzählen*.

Second, how does a dictionary discover the various meanings of words? One way is to analyze a corpus with concordancing, which gives the lexicographer exactly what is needed: many examples of a word in context. For example, Figure 6 shows that the phrasal verb *to add up* has both literal (lines 4 and 7) and figurative (lines 3 and 10) meanings.

### 3.2 Frequency Table Analyses
In the last section it was noted that word collocations are often words that appear together more often than chance. Clearly statistics can be helpful in quantifying this, which linguists have long known. Chapter 5 of *Foundations of Statistical Language Processing* (Manning and Schütze (1999)) mentions several techniques, including t-tests, chi-square tests, likelihood ratios and mutual information. The last method suggests that linguists have also delved into information theory, which is true. In this section, Fisher's exact test is applied to word pairs to check whether independence is a tenable hypothesis.

|  | **Word 1 present** | **Word 1 absent** | **Row sums** |
|---|---|---|---|
| **Word 2 present** | $c_{11}$ | $c_{12}$ | $c_{1\cdot}$ |
| **Word 2 absent** | $c_{21}$ | $c_{22}$ | $c_{2\cdot}$ |
| **Column sums** | $c_{\cdot 1}$ | $c_{\cdot 2}$ | $c_{\cdot\cdot}$ |

**Figure 7:** A frequency table for testing if two words are independent or not. If independence is rejected, then these two words may be a collocation.

Figure 7 shows a two-by-two contingency table to test whether or not two words are independent. If dependence is found, then it may be the case that the two words form a collocation. Using Program 6.1 (noted in the last section) applied to the Brown Corpus, two phrasal verbs are considered: *to throw up* vs. *to throw about*.

|             | **Throw** | **-Throw** | **Row Sums** |
|---:|---:|---:|---:|
| **Up**          | 8   | 1,966     | 1,974     |
| **-Up**         | 141 | 1,012,197 | 1,012,338 |
| **Column sums** | 149 | 1,014,163 | 1,014,312 |

**Figure 8:** A frequency table for testing if the words *throw* and *up* are independent or not. Since the phrasal verb *to throw up* is an idiom, one suspects dependence here. Note that the negative sign means the word after it does not appear.

For the native English speaker, the two phrases "throw up one's ball" and "throw up one's dinner" are easily understood. The first is not an example of a collocation since the meaning of the phrasal verb *to throw up* does follow from the individual meanings of the verb *to throw* and the preposition *up*. However, the second phrase is a collocation since here the phrasal verb means *to vomit*, which has nothing to do with *to throw*, and little to do with *up*.

To test the independence of the words *throw* and *up*, Fisher's exact test is performed. The p-value was computed using SAS's PROC FREQ with the result 7.92E-10, so these two words are dependent in the Brown Corpus. The expected value of $c_{11}$ is 0.29, and the p-value reveals that the observed value of 8 is significantly higher than that. With this example in mind, we consider Figure 9.

|             | **Throw** | **-Throw** | **Row Sums** |
|---:|---:|---:|---:|
| **About**       | 1   | 1,815     | 1,816     |
| **-About**      | 148 | 1,012,348 | 1,012,496 |
| **Column sums** | 149 | 1,014,163 | 1,014,312 |

**Figure 9:** A frequency table for testing to see if the words *throw* and *about* are independent or not.

For this second contingency table, the expected value of $c_{11}$ is 0.27, but now the observed value is 1, which is much lower than before. Not surprisingly, the p-value now is much larger, 0.2343, which suggests that the words *throw* and *about* are independent in the Brown Corpus. Unlike the phrasal verb *to throw up*, there is not an idiomatic use of *to throw about*, though it can be used both figuratively and literally.

## 4. Coupon Collecting and Pangrammatic Windows

This section has one last problem where language and statistics cross paths. A *pangram* is an English text that contains all 26 letters (ignoring case), and a pangrammatic window is a contiguous sample of text from a source that is also a pangram. The goal is to compare pangrammatic windows from Dickens' *A Christmas Carol* to what one would expect if the letters of the alphabet were independent of each other. Of course, letters are dependent, but it is interesting to see how independence may fail. We consider only one analysis here: comparing the length of the pangrams found in his novel compared to the lengths of pangrams via generating random letters using the empirical frequencies of the letters in his novel. Before starting this analysis, below is an example of a shorter-than-average pangram found in *A Christmas Carol*. Note that the letter *j* was the last to appear.

> The Spirit dropped beneath it, so that the extinguisher
> covered its whole form; but though Scrooge pressed it down
> with all his force, he could not hide the light: which streamed
> from under it, in an unbroken flood upon the ground.
>
> He was conscious of being exhausted, and overcome by an
> irresistible drowsiness; and, further, of being in his own
> bedroom. He gave the cap a parting squee**z**e, in which his hand
> relaxed; and had barely time to reel to bed, before he sank
> into a heavy sleep.
>
> AWAKING in the middle of a prodigiously tough snore, and
> sitting up in bed to get his thoughts together, Scrooge had
> no occasion to be told that the bell was again upon the
> stroke of One. He felt that he was restored to consciousness
> in the right nick of time, for the especial purpose of holding
> a conference with the second messenger dispatched to him
> through Jacob Marley's intervention.

## 4.1 The Coupon Collector's Problem and Comparing Pangram Lengths

The coupon collector's problem analyzes how long it takes to collect a full set of coupons, which are the 26 letters of the alphabet in this case. Of course, the letters of the alphabet do not appear equally often, but the coupon collector's problem has a closed-form solution when the coupons have unequal probabilities.

Deriving this solution is not easy, and the key result is merely quoted here. Let $p_i$ be the probability of the *i*th coupon, and let $N_{jk}$ be the number of coupons needed so that *j* distinct coupons each appear at least *k* times. Finally, let $e_k(t)$ be the *k*th order Taylor approximation of the exponential function, $e^t$. Then Theorem 2 of Flajolet et al. (1992) states the following.

$$E(N_{n,1}) = \int_0^\infty \left(1 - \prod_{i=1}^n (1 - \exp(-p_i t))\right) dt$$

Here *n* equals 26, the number of letters in the Roman alphabet. Although this integral is hard to do by hand, it is easily done with a symbolic mathematics package such as Mathematica. All that is needed are the estimates of the letter frequencies, $p_i$. This is easy to do with SAS: just read in the entire text one character at a time, then do a PROC FREQ for the letters *a* through *z*. The resulting counts and empirical frequencies are given in Figure 10.

| | | | | | | |
|---|---|---|---|---|---|---|
| a | 9308 | 0.076892 | | n | 7960 | 0.065756 |
| b | 1943 | 0.016051 | | o | 9690 | 0.080048 |
| c | 3035 | 0.025072 | | p | 2119 | 0.017505 |
| d | 5674 | 0.046872 | | q | 97 | 0.000801 |
| e | 14850 | 0.122674 | | r | 7031 | 0.058082 |
| f | 2433 | 0.020099 | | s | 7900 | 0.065261 |
| g | 2979 | 0.024609 | | t | 10869 | 0.089787 |
| h | 8368 | 0.069127 | | u | 3335 | 0.02755 |
| i | 8294 | 0.068515 | | v | 1022 | 0.008443 |
| j | 113 | 0.000933 | | w | 3096 | 0.025576 |
| k | 1031 | 0.008517 | | x | 131 | 0.001082 |
| l | 4553 | 0.037612 | | y | 2298 | 0.018983 |
| m | 2840 | 0.023461 | | z | 84 | 0.000694 |

**Figure 10:** A count and empirical frequency table for the letters of the alphabet in Dickens' novel, *A Christmas Carol*.

Using the Mathematica code given in Figure 11, the expected value of $N_{26,1}$ is 2473.82. Since the length of the quote at the beginning of Section 4 is 680 letters (this ignores spaces and punctuation), it is much shorter than average.

```
table = Import["c:\christmascarol.csv"]; (* Contains Figure 10 data *)
p = table[[All,3]];
NIntegrate[1-Product[1-Exp[-p[[i]] t], {i,1,Length[p]}], {t,0,Infinity}]
```

**Figure 11:** Computing $E(N_{26,1})$ using Mathematica.

To finish this analysis, a thousand pangrammatic windows were found in *A Christmas Carol* by picking paragraphs at random then collecting text until all letters were found. Then a thousand additional pangrams were created by generating random letters using the proportions given in Figure 10. Before looking at the histograms of the lengths of these pangrams on the next page (Figures 12 and 13), how similar should these histograms be? Since writers do not use letters at random, one expects some differences, perhaps great dissimilarities.

It turns out that the two histograms are similar. Both have modes just above 1500 letters. In fact, their shapes are alike up to about 7,500. However, Figure 12 has a long right tail, while Figure 13 stops abruptly at 7,500. That is, *A Christmas Carol* has much longer stretches where there is at least one letter missing.

The longer tail in Figure 12 is mostly due to the following. It is caused by a rare letter not appearing as often as it would given independence. It turns out that of the 84 *z*s in *A Christmas Carol*, the name Fezziwig appears 20 times accounting for 40 of these, all of which appear on just three pages in Stave 2. Hence there are only 44 *z*s to be distributed through out the rest of the novel, which is roughly half the rate of the *z*s for Figure 13.

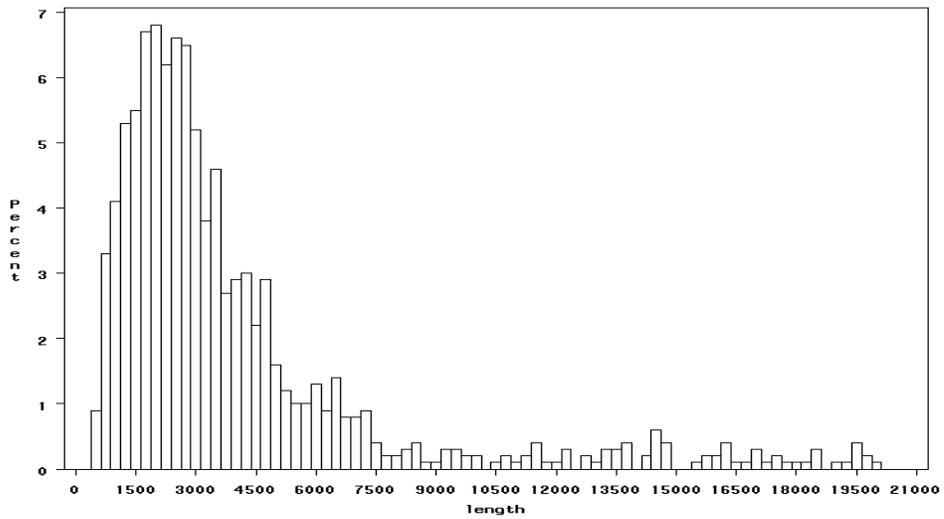

**Figure 12:** Histogram of the lengths of 1000 pangrammatic windows found by picking random starting places in Dickens' *A Christmas Carol*. Note that both this and the next histogram are drawn to the same scale.

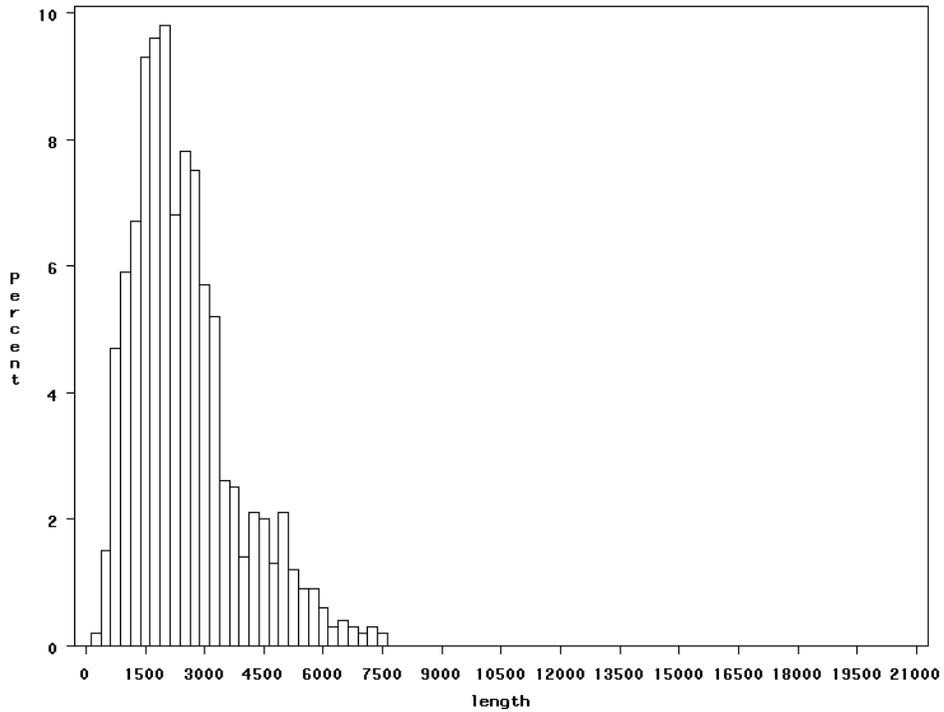

**Figure 13:** Histogram of the lengths of 1000 pangrams found by generating letters independently using the letter proportions given in Figure 10, which are the letter proportions found in *A Christmas Carol*.

## 4.2 A Birthday Problem Aside

Flajolet et al. (1992) give a general formula for $E(N_{jk})$ and point out how the birthday problem is related to $N_{12}$. If one looks at actual birthday data for a specific year (Chance (2009) gives U.S. data for 1978), it turns out that birthdays are **not** distributed uniformly because there a more than 10% drop on weekends and holidays along with a smaller seasonal affect as shown in Figure 14. It turns out that this barely changes $E(N_{12})$: assuming uniformity it is 24.62, but using the 1978 data it drops to only 24.53. However, Figure 14 makes a great example for an introductory statistics class because it generates discussion.

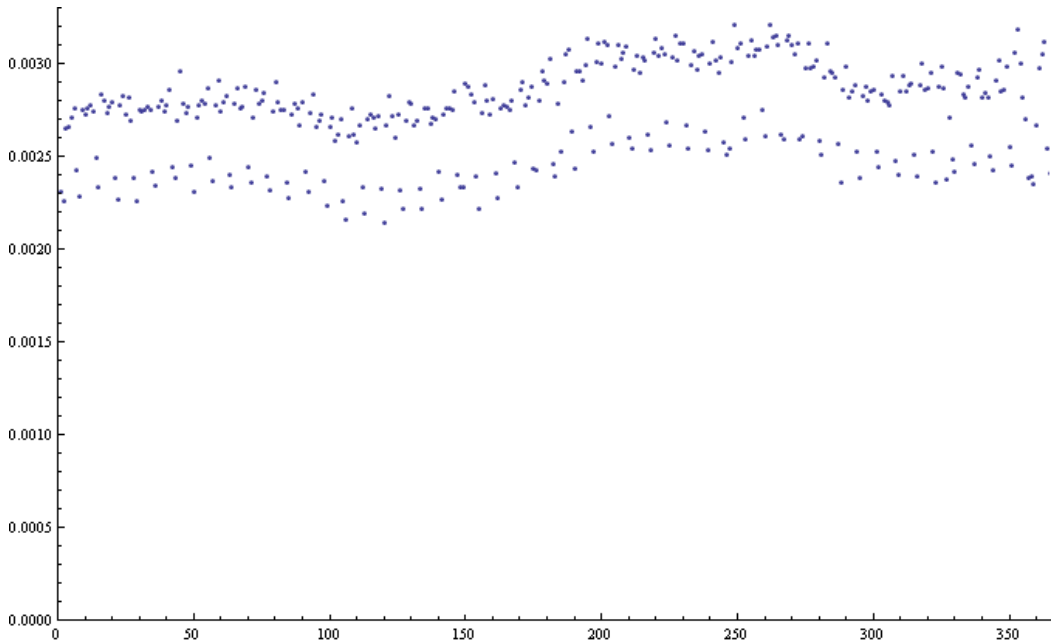

**Figure 14:** Plot of the number of U.S. birthdays in 1978 vs. day of the year (that is, 1 = Jan. 1st, …, 365 = Dec. 31st). The lower band is due to weekends and holidays. This data is available from http://www.dartmouth.edu/~chance/teaching_aids/data/birthday.txt (see Chance (2009) for details.)

## 5. Conclusions

There are two lessons to be learned from this paper. First, the linguists already have created many statistical examples, which is a great place to start for a statistics teacher wanting to include examples using language. Of especial interest are the corpus linguists who believe in creating samples of text that are representative of some aspect of language, and then employ computers to do numerous analyses. Second, statistical examples using language can be used in a variety of statistics courses including statistical programming, categorical data analysis, multivariate data analysis, and applied statistics.

I have just starting incorporating language examples in my classes with generally positive results. However, note that some non-native speakers of English can find such examples difficult to understand.

## Acknowledgements

Thanks to my STAT 456 class (Introduction to SAS Programming for spring semester, 2009) for letting me try out some language examples for the first time. I owe a giant debt to all the people like Grady Ward who have released language data to the public domain and to all the sites like Project Gutenberg that provide public domain texts.